\documentclass{article}

\usepackage[nonatbib, final]{ml4mols_2020}

\usepackage[utf8]{inputenc} % allow utf-8 input
\usepackage[T1]{fontenc}    % use 8-bit T1 fonts
\usepackage{hyperref}       % hyperlinks
\usepackage{url}            % simple URL typesetting
\usepackage{booktabs}       % professional-quality tables
\usepackage{amsfonts}       % blackboard math symbols
\usepackage{nicefrac}       % compact symbols for 1/2, etc.
\usepackage{microtype}      % microtypography
\usepackage{cite}
\usepackage{graphicx}
\usepackage{subcaption}

\title{Gini in a Bottleneck:  Sparse Molecular Representations for Graph Convolutional Neural Networks}

\author{
Ryan ~Henderson \\
\texttt{ryan.henderson@bayer.com} \\
\And
Djork-Arn\'{e} ~Clevert\\
\texttt{djork-arne.clevert@bayer.com} \\
\And
Floriane ~Montanari \\
\texttt{floriane.montanari@bayer.com} \\
\\
Digital Technologies \\
Bayer AG \\
13353 Berlin, Germany \\
}

\begin{document}

    \maketitle

    \begin{abstract}
        Due to the nature of deep learning approaches, it is inherently difficult to understand which aspects of a molecular graph drive the predictions of the network.
        As a mitigation strategy, we constrain certain weights in a multi-task graph convolutional neural network according to the Gini index to maximize the ``inequality'' of the learned representations.
        We show that this constraint does not degrade evaluation metrics for some targets, and allows us to combine the outputs of the graph convolutional operation in a visually interpretable way.
        We then perform a proof-of-concept experiment on quantum chemistry targets on the public QM9 dataset, and a larger experiment on ADMET targets on proprietary drug-like molecules.
        Since a benchmark of explainability in the latter case is difficult, we informally surveyed medicinal chemists within our organization to check for agreement between regions of the molecule they and the model identified as relevant to the properties in question.
    \end{abstract}

    \section{Introduction}

    Multi-task graph convolutional neural networks (GCNs) have become competitive in predicting some molecular properties e.g. for drug screening~\cite{feinberg_potentialnet_2018, montanari_modeling_2019}.
    However, their opaque nature remains a stumbling block for wider adoption within the chemistry community.
    While several approaches exist for rationalizing the predictions of GCNs applied to molecular problems ~\cite{baldassarre_explainability_2019, preuer_interpretable_2019, ying_gnnexplainer_2019}, we take a somewhat orthogonal approach and introduce a bottleneck in the training itself.
    % For practical reasons, most of all data imbalance, we favor multitask GCNs~\cite{montanari_modeling_2019}.

\begin{figure}[h]
        \centering
        \begin{subfigure}{.5\textwidth}
            \centering
            \includegraphics{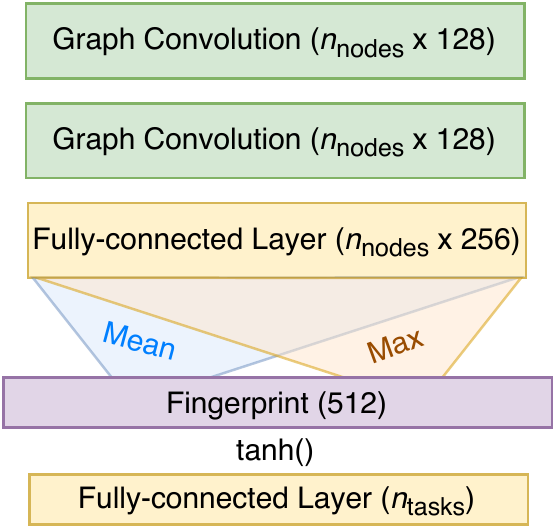}
            \caption{Reference architecture}\label{fig:reference_architecture}
        \end{subfigure}%
        \begin{subfigure}{.5\textwidth}
            \centering
            \includegraphics{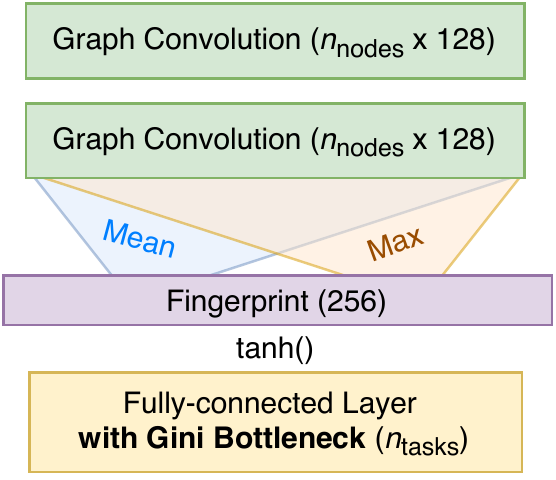}
            \caption{Explainable architecture}\label{fig:explainable_architecture}
        \end{subfigure}
        \caption{
        Multi-task GCN reference and explainable architectures.
        Node-level aggregations are gathered from the graph embedding to produce the fingerprint.
        We remove the intermediate fully-connected layer from the reference architecture so that the components of a prediction can be easily linked to node-level representation.
        The Gini regularization is applied to the weights of the final linear layer.
        The size of the output of each layer is in parenthesis.
        Each layer before the fingerprint is followed by a batch normalization layer and ReLU activation (not pictured).
        }
        \label{fig:architecture}
    \end{figure}

    A typical architecture (see Figure~\ref{fig:reference_architecture}) converts the final node-level aggregations or ``fingerprint'' into the chemical targets of interest with a linear layer.
    In this work, we introduce a regularizer on the weights of this layer to enforce sparsity in the hopes that this will reduce the number of relevant node-level aggregations necessary to inspect to explain a model's prediction.
    The usual $\ell_1$ or $\ell_2$ regularization is not appropriate here, as penalizing the magnitude of the weights directly damages the performance of regression metrics.
    Instead, we borrow from economics and use the Gini coefficient~\cite{gini_variabilita_1912, noauthor_gini_2020} as a regularizer:

    \begin{equation}
        g = \sum_i^n \sum_j^n \frac{
        | w_i - w_j |
        } {
        2 n^2 \bar{w}
        }
        \label{eq:gini}
    \end{equation}

    where $i,j$ range over all weights in the layer.
    The Gini coefficient ranges from zero to one: zero if all $w$ are equal and one if one $w_i$ is non-zero and the rest zero.
    Since weights of a linear transform are not necessarily restricted to be non-negative, we will always use $|w|$ rather than $w$ in our calculations.
    The training loss becomes $L / g^m$ where $L$ is the multi-task regression loss, and $m$ is a hyperparameter to tune the effect of the Gini regularization.
    These changes are illustrated in Figure~\ref{fig:explainable_architecture}.

    \section{Experiment Setup}

    \paragraph{Graph Convolutional Neural Network}

    We use molecular fingerprint convolutions~\cite{duvenaud_convolutional_2015} in all our network architectures and the QM9 dataset~\cite{ramakrishnan_quantum_2014, ruddigkeit_enumeration_2012} for training.
    We use the GCN multitask architecture given by~\cite{montanari_modeling_2019} but with multitask targets adjusted for the QM9 dataset and a smaller featurization given in~\cite{gilmer_neural_2017}.
    Since max and mean aggregations are applied to the node representations and concatenated to generate the molecular fingerprint, we apply separate Gini regularizers to weights of the output layer which operate on each aggregation.
    Though we are benchmarking against the Deepchem implementation~\cite{wu_moleculenet_2018, ramsundar_deep_2019}, we have implemented everything with Pytorch Geometric~\cite{fey_fast_2019}.

    \paragraph{Quantum Electronic Structure Calculations}

    To verify that the representations selected by our model have chemical relevance, we wish to compare them to some visual representation on the molecule.
    We chose the highest- and lowest-occupied molecular orbitals (HOMO and LUMO) as their energies are included as targets in QM9.
    Since the HOMO and LUMO electron densities have more complex spatial structures than node representations generated by molecular convolutions, which are limited to one real number per atomic site, we opt to compute the condensed Fukui function\cite{parr_density-functional_1994, yang_use_1986}.
    The Fukui function gives the change in electronic population at the $k$ site for the removal or addition of one electron to the molecule:
    \begin{equation}
        f^-_k = \rho_k(N) - \rho_k(N-1)
        \label{eq:fukui_remove}
    \end{equation}
    \begin{equation}
        f^+_k = \rho_k(N+1) - \rho_k(N)
        \label{eq:fukui_add}
    \end{equation}

    The $f^-_k$ and $f^+_k$ can be regarded as the relative nucleophilicity and electrophilicity of each site, respectively.
    Intuitively, sites with high nucleophilicity $f^-_k$ correspond to electron population in the HOMO, whereas sites with high electrophilicity $f^+_k$ correspond to electron population in the LUMO (if an electron were added to the molecule).
    We propose, therefore, that \textit{interpretable learned convolutional representations of the HOMO and LUMO will correspond to $f^-_k$ and $f^+_k$}, respectively.
    The \texttt{xtb}~\cite{bannwarth_extended_2020} package is used to calculate the Fukui functions.
    % For instance, if the coordinates from the QM9 dataset are in \texttt{coords.xyz}, then we would execute: \texttt{xtb coords.xyz ----opt \&\& xtb xtbopt.xyz ----vfukui}

    \section{Results}

\begin{figure}
        \centering
        \begin{subfigure}{.5\textwidth}
            \centering
            \includegraphics[width=\linewidth]{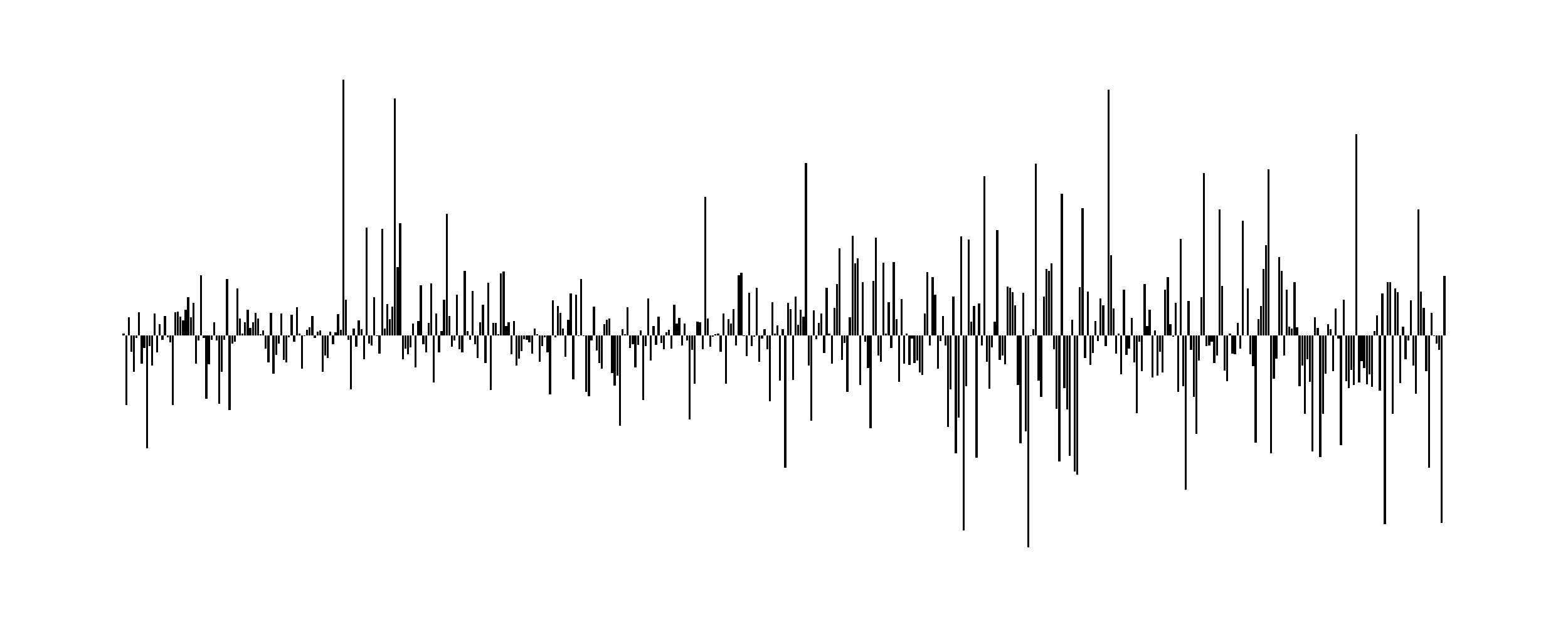}
            \caption{Baseline (corresponds to Figure~\ref{fig:reference_architecture})}\label{fig:default_weights}
        \end{subfigure}%
        \begin{subfigure}{.5\textwidth}
            \centering
            \includegraphics[width=\linewidth]{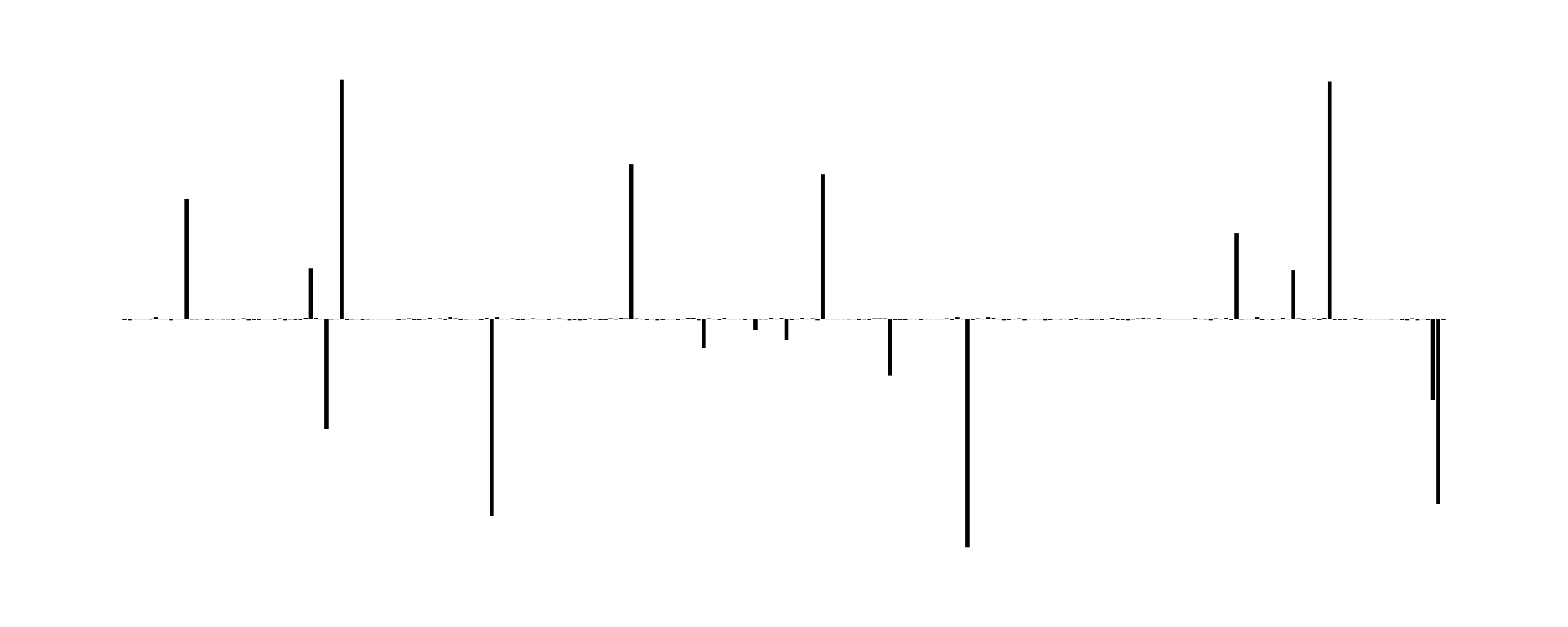}
            \caption{Explainable (corresponds to Figure~\ref{fig:explainable_architecture})}\label{fig:explainable_weights}
        \end{subfigure}
        \caption{
            Illustration of weight values in final fully-connected layer for the column corresponding to the prediction of the HOMO energy.
            In the Gini-constrained explainable architecture, we can see that only about a dozen learned representations are relevant to this endpoint.
        }
        \label{fig:HOMOweights}
    \end{figure}

    Our reported results (Table \ref{tab:results}) are the mean of random 5-fold cross-validation.
    HOMO and LUMO energy errors are stable across architectures, while other targets suffer some performance degradation from the multi-task setting and introduction of the Gini bottleneck.
    The effect of the Gini bottleneck on the weights of the final linear layer is depicted in Figure~\ref{fig:HOMOweights}.

    \begin{table}
        \centering
        \caption{
            Comparison of mean absolute error (MAE) of different architectures.
            The MTGCN (multi-task graph convolutional network) explainable model uses the architecture depicted in Figure~\ref{fig:explainable_architecture} and Gini factor $m=10$.
            Best values are bold.
            Explanations of each target is found in~\cite{ramakrishnan_quantum_2014}.
            Additional targets $U,H,G$ are also computed but not shown, having very similar results as $U_0$.}
        \begin{tabular}{llrrr}
            \toprule
            Target  & Units & MTGCN Baseline    & MTGCN Explainable & Single task GCN~\cite{wu_moleculenet_2018} \\
                    &  & (this work)    &  (this work)  &  \\
            \midrule
            $\mu$   &  Debye & 0.615            & 0.604             & \textbf{0.583}                            \\
            $\alpha$&  Bohr$^3$ & 1.961            & 2.567             & \textbf{1.370}                            \\
            HOMO    &  Hartree & 0.00609         & \textbf{0.00606} & 0.00716                                   \\
            LUMO    &  Hartree & \textbf{0.00681} & 0.00716           & 0.00921                                   \\
            delta   &  Hartree & \textbf{0.00896} & 0.00913           & 0.01120                                   \\
            R2      &  Bohr$^2$ & 81.8             & 95.4              & \textbf{35.9}                             \\
            ZPVE    &  Hartree & 0.00409          & 0.00538           & \textbf{0.00299}                          \\
            $U_0$   &  Hartree & 10.544           & 13.077            & \textbf{3.410}                            \\
%            $U$    &   & 10.543911         & 13.305641         & \textbf{3.41000}  \\
%            $H$    &   & 10.547448         & 13.199469         & \textbf{3.41000}  \\
%            $G$    &   & 10.540622         & 13.373771         & \textbf{3.41000}  \\
            $C_v$   &  cal/mol$K$ & 0.846            & 1.112             & \textbf{0.650}                            \\
            \bottomrule
        \end{tabular}\label{tab:results}
    \end{table}

    \paragraph{HOMO and LUMO}

\begin{figure}[h]
        \centering
        \begin{subfigure}{.30\textwidth}
            \centering
            \includegraphics[width=\linewidth]{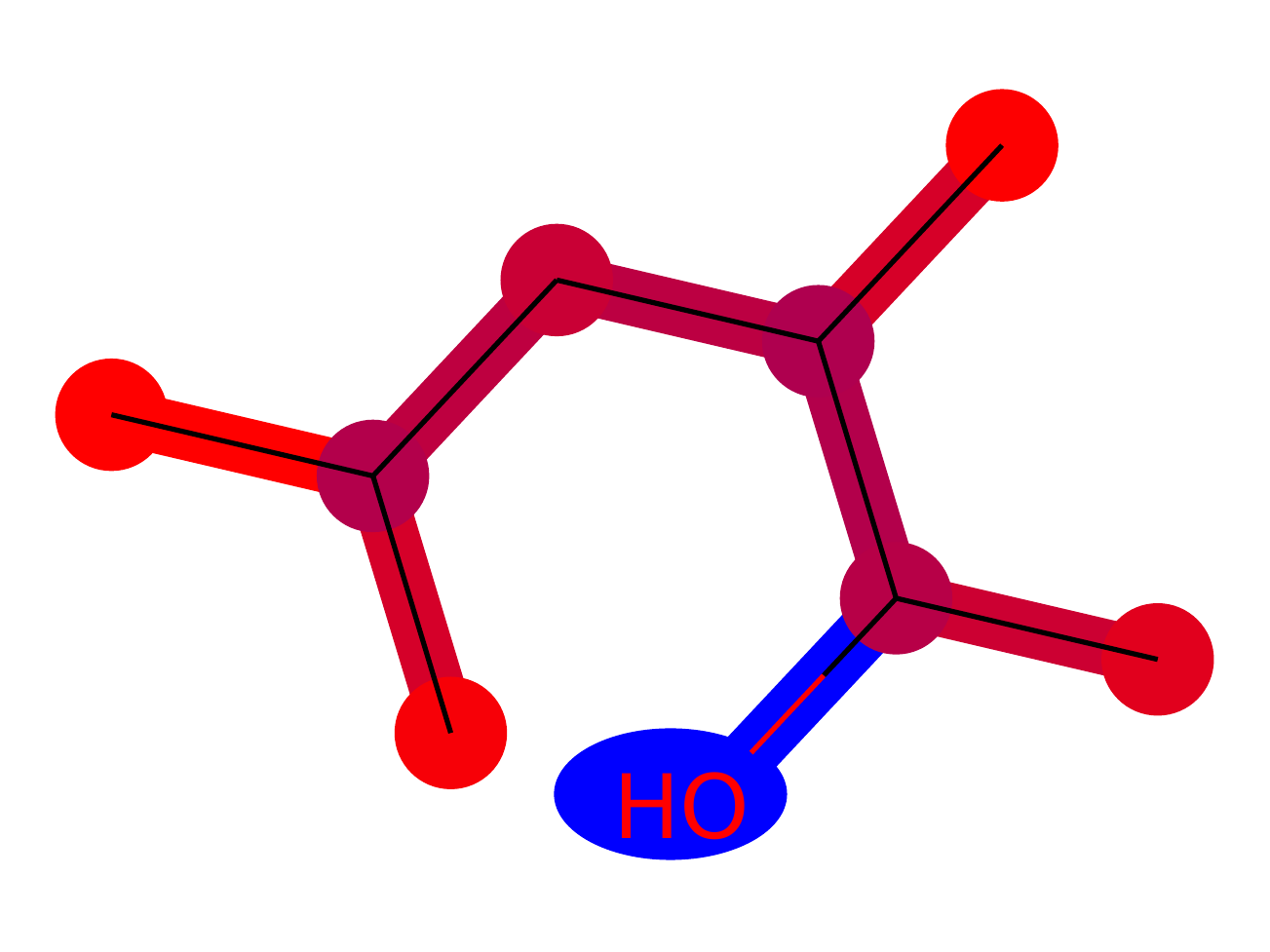}
            \caption{$f^-_k$}
        \end{subfigure}%
        \begin{subfigure}{.30\textwidth}
            \centering
            \includegraphics[width=\linewidth]{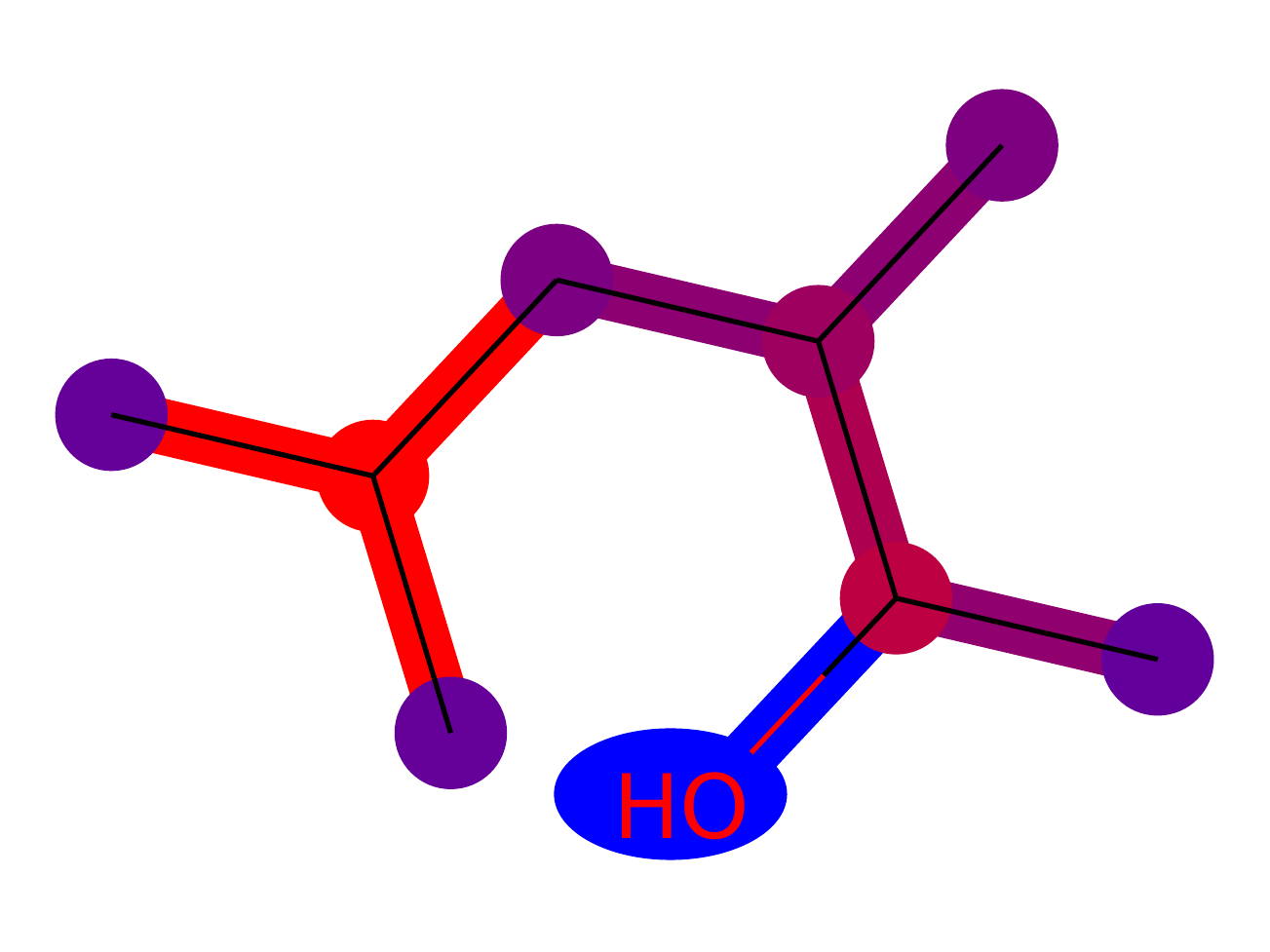}
            \caption{HOMO Contribution}
        \end{subfigure}%
        \begin{subfigure}{.40\textwidth}
            \centering
            \includegraphics[width=\linewidth]{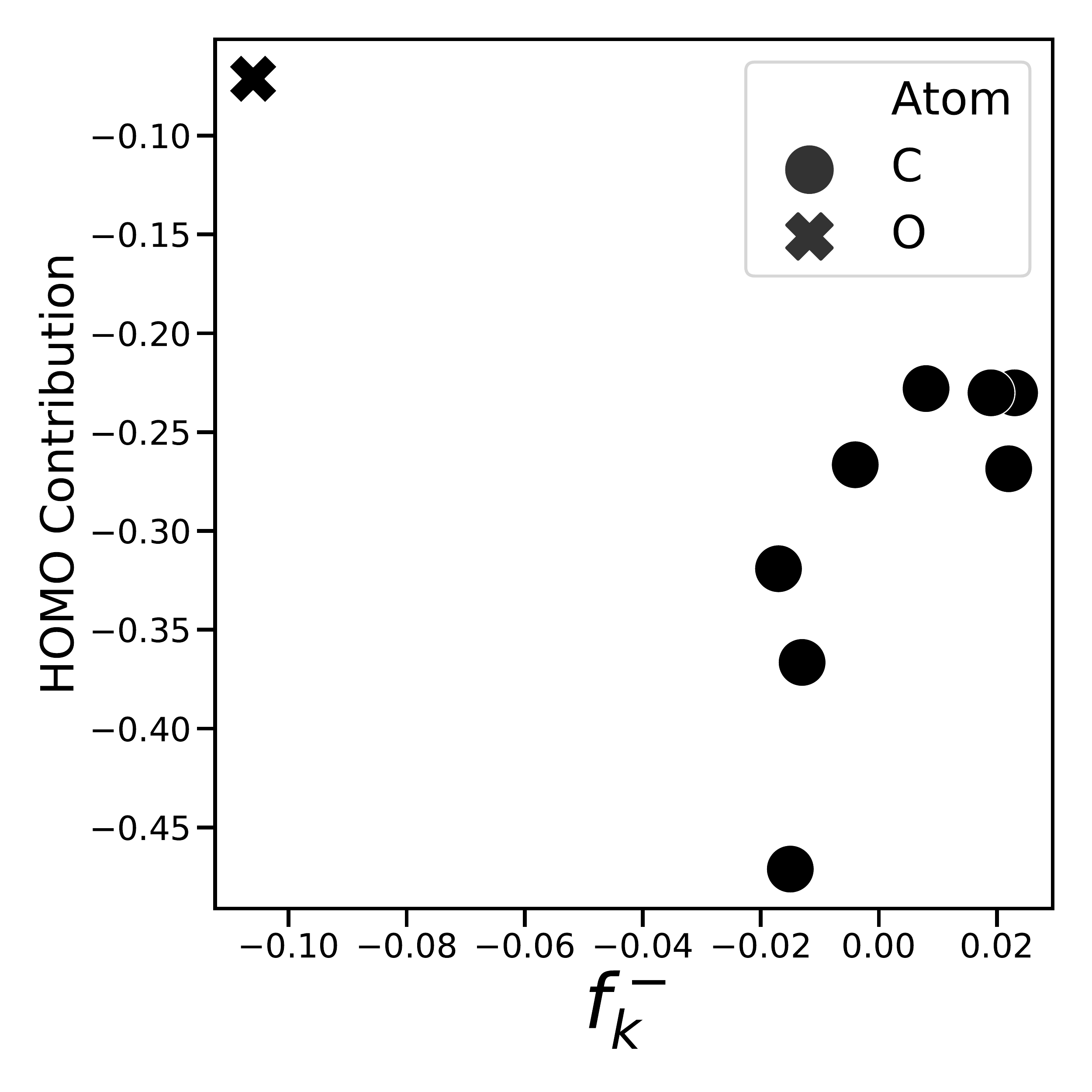}
        \end{subfigure}
        \caption{
        $f^-_k$ (Equation~\ref{eq:fukui_remove}) and the unscaled contribution to the predicted HOMO energy for each site from the explainable GCN.
        In the accompanying plot, the hydroxyl group is isolated with the lowest $f^-_k$ value and least negative contribution to the HOMO energy prediction.
        }
        \label{fig:fuk_vs_exp}
    \end{figure}

    Since we have removed the intermediate fully-connected layer in the explainable architecture (Figure~\ref{fig:explainable_architecture}), each weight (Figure~\ref{fig:explainable_weights}) corresponds to the mean or max of a learned convolutional node-level representation.
    These representations are sparse enough to be considered individually, but for brevity's sake we will combine them into one picture as a contribution-weighted linear combination of all node outputs (mean) or maximum node output (max) from the last graph convolutional layer.
    The contribution to the final prediction is $w_{ij} \tanh(f(x_i))$, where $w_{ij}$ is a weight in the last fully-connected layer corresponding to the  $j$\textsuperscript{th} endpoint and $i$\textsuperscript{th} node representation, $x_i$.
    $f$ is max or mean.
    Figure~\ref{fig:fuk_vs_exp} shows how the contribution-weighted sum of the node representations compare with the Fukui function for an example molecule from the QM9 dataset.

    \paragraph{Survey of Medicinal chemists}

\begin{figure}
        \begin{subfigure}{0.5\textwidth}
            \centering
            \includegraphics[width=0.6\linewidth, angle=90, trim=210 385 170 110, clip]{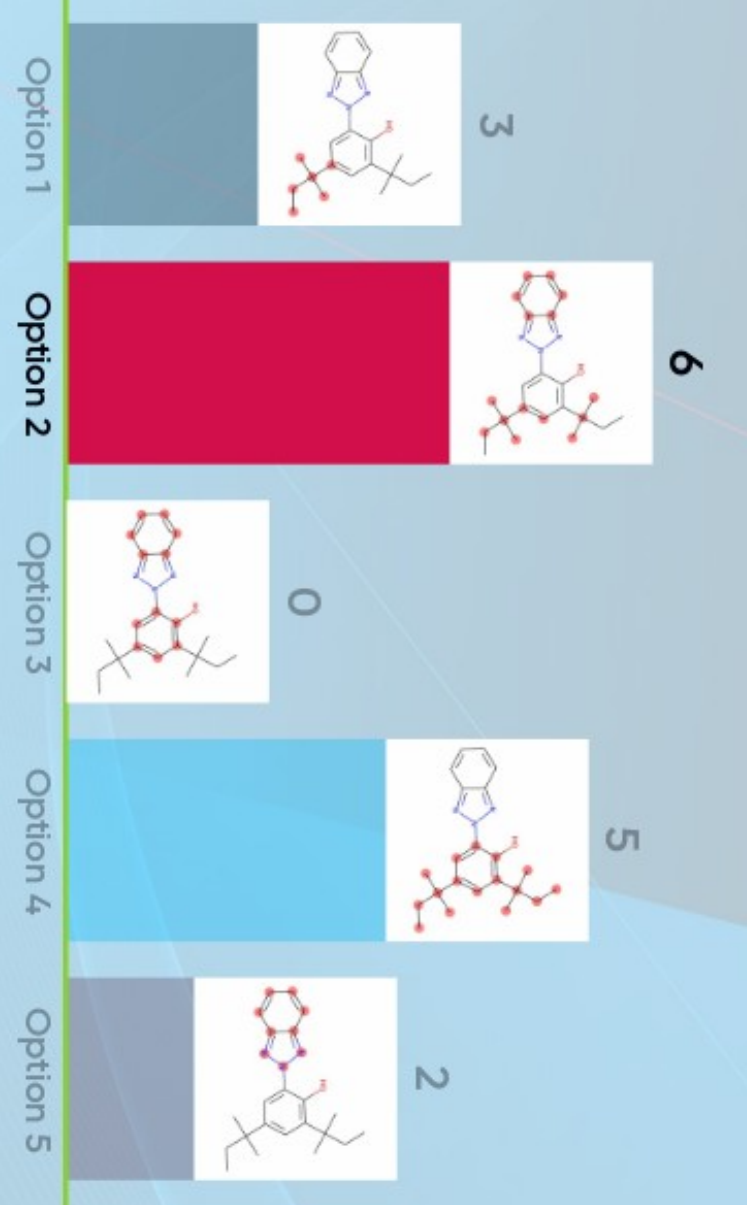}
            \caption{Example question}
            \label{fig:example_question}
        \end{subfigure}%
        \begin{subfigure}{0.5\textwidth}
            \caption{Summary of responses}\label{tab:game_results}
                \begin{tabular}{lrrrr}
                    \toprule
                    {}                & $n$ & $\mu$ & Top 1 & Top 2 \\
                    \midrule
                    HSA Binding       & 6   & 17.3 & 0.67  & 1.00  \\
                    Melting Point     & 2   & 15.5 & 0.00  & 0.50  \\
                    Membrane affinity & 3   & 9.0 & 0.33  & 1.00  \\
                    Solubility        & 10  & 18.7 & 0.30  & 0.70  \\
                    \bottomrule
                \end{tabular}
        \end{subfigure}
        \caption{
            ~\subref{fig:example_question})
            Example solubility question, screenshot from presentation.
            In this case, a plurality of chemists agreed with the model (Option 2).
            Note the high variation within the responses.
            ~\subref{tab:game_results})
            Fraction of questions where explanation proposed by our model was in the top or top two most chosen responses.
            $n$ questions from each endpoint were posed, with an average of $\mu$ responses per question within each property.
            HSA is short for human serum albium.
            Further explanation of the targets can be found in~\cite{montanari_modeling_2019}.
        }
    \end{figure}

    While the results on the QM9 dataset are encouraging, the molecules are not drug-like and the given targets reflect that.
    We retrained our model on a corpus of drug-like molecules~\cite{montanari_modeling_2019} with measured targets of interest to medicinal chemists~\cite{kramer_learning_2018}.
    Since the model had reasonable performance (similar to what is reported in~\cite{montanari_modeling_2019}), we were confident we could generate good rationalizations of the various properties as in the previous section.
    We presented a few unseen examples to chemists within our organization in an interactive presentation with atomic contributions highlighted as in Figure~\ref{fig:fuk_vs_exp}, along with a few random but plausible alternatives.
    See Figure~\ref{fig:example_question} for an example question.
    Agreement between the audience and the explainable model is summarized in Table \ref{tab:game_results}.

%    \begin{figure}
%        \centering
%        \includegraphics[width=0.6\linewidth, angle=90, trim=210 385 170 110, clip]{figures/answer_1.pdf}
%        \caption{
%        Example solubility question.
%        In this case, the a plurality of chemists agreed with the model.
%        Note the high variation within the responses.
%        }
%        \label{fig:example_question}
%    \end{figure}
%
%    \begin{table}
%        \centering
%        \caption{
%        Fraction of questions where model was in the top or top two most chosen responses.
%        $n$ questions from each endpoint were posed.
%        }
%        \begin{tabular}{lrrr}
%            \toprule
%            {}                & $n$ & Top 1 & Top 2 \\
%            \midrule
%            HSA Binding       & 6   & 0.67  & 1.00  \\
%            Melting Point     & 2   & 0.00  & 0.50  \\
%            Membrane affinity & 3   & 0.33  & 1.00  \\
%            Solubility        & 10  & 0.30  & 0.70  \\
%            \bottomrule
%        \end{tabular}\label{tab:game_results}
%    \end{table}

    \section{Conclusions \& Future Work}

    We demonstrated a new method for reducing the number of relevant learned representations for a regression target in a multi-task graph neural network.
    We further show that this constraint can be applied in some cases without degrading the model performance, and that the remaining representations are qualitatively connected to the predicted property.

    In the future we want to combine this sparsity with orthogonalization techniques to make sure the relevant learned representations are not only few, but mutually informative.
    We will further explore how best to combine the representations into actionable rationalizations,  and continue to query practicing chemists on the usefulness of the generated rationalizations.

    \begin{ack}
        We would like to thank Andreas Goeller for suggesting the use of Fukui functions.
        Funding in direct support of this work: Bayer AG Life Science Collaboration (``Explainable AI'').
    \end{ack}

    \medskip

    \small
    \bibliography{neurips_2020}{}
    \bibliographystyle{plain}
\end{document}